\documentclass{Interspeech2024}
\usepackage{multirow}
\usepackage{graphicx}
\usepackage{pifont}
\usepackage[table,xcdraw]{xcolor}

\interspeechcameraready

\title{MultiPA: A Multi-task Speech Pronunciation Assessment Model for Open Response Scenarios}

\name{Yu-Wen}{Chen}
\name{Zhou}{Yu}
\name{Julia}{Hirschberg}

\address{
  Department of Computer Science, Columbia University, United States}
\email{\{yuwchen, zhouyu, julia\}@cs.columbia.edu}


\begin{document}

\maketitle

\begin{abstract}
Pronunciation assessment models designed for open response scenarios enable users to practice language skills in a manner similar to real-life communication. However, previous open-response pronunciation assessment models have predominantly focused on a single pronunciation task, such as sentence-level accuracy, rather than offering a comprehensive assessment in various aspects. We propose MultiPA, a Multitask Pronunciation Assessment model that provides sentence-level accuracy, fluency, prosody, and word-level accuracy assessment for open responses. We examined the correlation between different pronunciation tasks and showed the benefits of multi-task learning. Our model reached the state-of-the-art performance on existing in-domain data sets and effectively generalized to an out-of-domain dataset that we newly collected. The experimental results demonstrate the practical utility of our model in real-world applications.

\end{abstract}

\section{Introduction}

An automatic speech pronunciation assessment model offers a less expensive and more efficient approach to practicing and evaluating pronunciation skills in a second language (L2) \cite{ehsani1998speech, egan1999speaking}. One common design of the pronunciation assessment model is the closed-response scenario. In the closed-response scenario, L2 learners are instructed to speak a predetermined target sentence, which is then used as the ground-truth transcript for the model. However, this design restricts evaluation to predefined sentences and fails to reflect learners' pronunciation skills in real-world communication. In contrast, the open-response scenario allows learners to  speak freely or to respond to a given task or question, providing a more authentic evaluation of learners' pronunciation skills. Therefore, in this study, we aim to develop a pronunciation model that can assess learners' pronunciation skills in this open response scenario. 

One of the most commonly used methods for pronunciation assessment is using Goodness of Pronunciation (GoP) features~\cite{gong2022transformer, liu2023leveraging, sheoran2023pronunciation, pei2023gradformer, chao2023hierarchical, shekar2023assessment}. These features are calculated from the posterior probability of the ground-truth phone using an automatic speech recognition (ASR) model~\cite{hu2015improved}. Therefore, GoP-based models are designed and evaluated under the assumption of having accurate transcripts. As a result, their performance may experience a notable decline when directly applied to open-response scenarios where alignment between audio and accurate transcripts is not available~\cite{liang2023end}. Without access to a target sentence or ground-truth transcript, testing in the open response scenario requires a non-intrusive assessment model. Such a model can either utilize the ASR recognition results as its transcript or operate without using a transcript for evaluation~\cite{zhang2021end, zezario2022mti, chen22i_interspeech}. However, studies on non-intrusive pronunciation assessment have predominantly centered on a single pronunciation task~\cite{chung2017spoken, mao2019nn, fontan2020using}. For example,~\cite{lin2021deep} employed deep features from a DNN-HMM-based acoustic model to obtain a sentence-level total score.~\cite{liu2023zero} proposed a self-supervised learning (SSL)-based zero-shot model to assess sentence-level total proficiency.~\cite{lin2023exploiting} fused acoustic and phoneme representations to obtain sentence-level accuracy scores.~\cite{liu2023asr} developed an SSL-based ASR-free approach for fluency assessment. None of these previous studies have explored non-intrusive multitask pronunciation assessment for both sentence-level and word-level scores. In a multi-task assessment model, sentence-level accuracy, fluency, and prosody assessments provide L2 learners with a comprehensive overview of their pronunciation skills, while word-level scores pinpoint specific parts of their speech that require practice.

In this study, we propose MultiPA, a Multi-task Pronunciation Assessment model for open-response scenarios. Compared with the previous non-intrusive pronunciation assessment models, MultiPA provides a more comprehensive pronunciation assessment, including both sentence- and word-level assessments. Furthermore, we are the first to conduct a pilot study to evaluate the model's performance in real-world open-response scenarios. To do this, we collected data from L2 learners who were using an English learning chatbot to practice English and recruited experts for the annotation. The experimental results show the effectiveness of using our model in the real-world use case, and the pilot data we collected will be released for other studies to evaluate their model on multitask pronunciation assessment. 

\section{Our Multi-task Pronunciation Assessment model}
\begin{figure*}
\centering
\centerline{\includegraphics[scale=0.89]{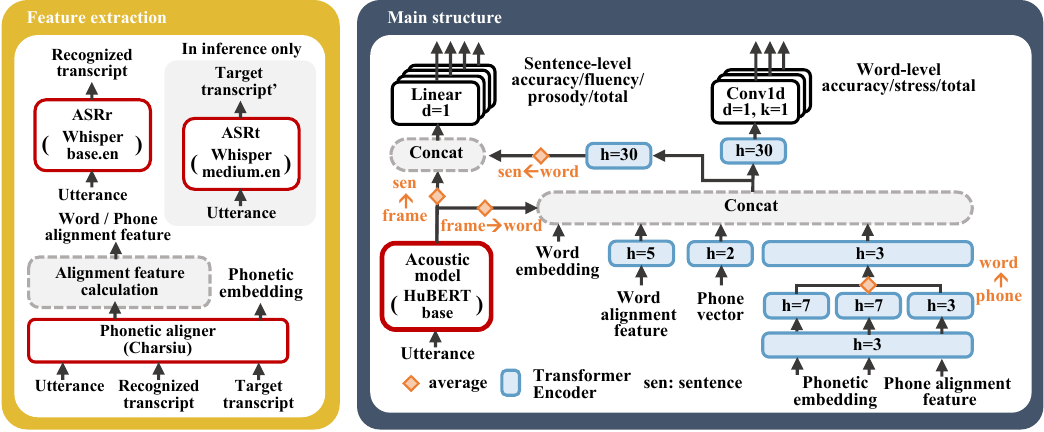}}
\caption{Overview of MultiPA, where $d$ in Linear and Conv1d layers refers to the output dimension, $k$ is the kernel size, and $h$ indicates the number of heads. The selection of $h$ is based on empirical results.}
\label{fig:overview}%
\end{figure*}

MultiPA utilizes a pretrained SSL model (i.e., HuBERT~\cite{hsu2021hubert}) as its main structure. Fine-tuning pre-trained SSL models has proven effective in phone-level mispronunciation detection~\cite{xu2021explore}, phone-level assessment~\cite{zahran2023fine}, and sentence-level assessment~\cite{kim2022automatic, fu2023phonetic}, but has not been used for assessing multi-task sentence-level and word-level scores. To provide word-level evaluation, it is necessary to identify the boundaries of individual words within the speech signal. MultiPA achieves this by first using the ASR model Whisper~\cite{radford2023robust} to identify potential words in the speech signal, followed by using Charsiu~\cite{zhu2022charsiu} to obtain alignment information between the words and the speech signals. Given the potential inaccuracies of the ASR, RoBERTa~\cite{liu2019roberta} was employed to offer supplementary semantic feedback for the recognized words. Figure~\ref{fig:overview} shows an overview of MultiPA. 

\subsection{Auditory feature extraction}
MultiPA extracts features from two transcripts: the target transcript and the recognized transcript. The target transcript represents the sentence that the learner wants to say, while the recognized transcript is how an ASR model (ASRr) recognized the speech signal. Although the target transcript is unavailable during inference, we still use it during training because word-level scores of training data are aligned with the target transcript. During inference, we use the recognition results of another ASR model (ASRt) as an alternative to the target transcript. Specifically, Whisper base.en (ASRr) generates the recognized transcript, while a larger ASR model, Whisper medium.en (ASRt), replaces the target transcript. Then, the phonetic aligner Charsiu provides word and phone alignment information between the transcript and the utterance. These aligned transcripts enable the extraction of word-level and phone-level features.

\subsubsection{Word-level features}
Word-level features are aligned on a word-by-word basis, with the length equal to the number of words in the target transcript. These features include word-embeddings, phone vectors, and word-alignment features. The word-embedding is the concatenation of RoBERTa~\cite{liu2019roberta} embeddings from target and recognized transcripts. The phone vector is the one-hot-encoded phones of the word. The word-alignment feature is computed using the alignment information with the transcripts, including duration (measuring the time each word takes), interval (indicating the time gap from the preceding word), time difference (capturing the variance in start and end times between words in the two transcripts), distance (reflecting the Levenshtein distance between words in the transcripts), aligned word count (counting the number of words aligned with the target word), phone distance (quantifying the matched phones between target word and aligned recognized words), and phone ratio (expressing the ratio of phones in the target word to those in the recognized word).

\subsubsection{Phone-level features}
Phone-level features are aligned on a phone-by-phone basis with the length equal to the number of phones in the target transcript. The phone-level features contain phonetic embedding and phone alignment features. The phonetic embedding is the output layer of the Charsiu, whose value indicates the probability of all possible phones. Phone alignment features include duration, interval, time difference, aligned phone count, and phone probability, where duration, interval, time difference, and aligned phone count have a similar definition as the word alignment features but are calculated on a phone basis. Lastly, the phone probability is the probability of aligning the specific targeted phone or recognized phone to the signals. 

\subsection{Main structure}
The main structure of MultiPA is based on fine-tuning a pretrained self-supervised-learning (SSL)-based model, HuBERT, with additional layers. MultiPA employs transformerEncoder layers for feature fusion and uses average pooling and alignment information to align features at different levels (e.g., aligning phone-level features to word-level features). Finally, linear layers and convolutional layers are used for sentence- and word-level assessment, respectively.

\section{Experimental setup}

\subsection{Data}
We utilize two datasets in our study: the open-source pronunciation assessment dataset speechocean762~\cite{speechocean762} and our self-collected data (referred to as multiPA data). The speechocean762 dataset serves as both the training and in-domain testing, while the multiPA dataset is used for out-of-domain testing. The speechocean762 dataset was collected in a closed-response scenario with known ground-truth transcripts. However, to simulate an open-response scenario, we did not use the ground-truth transcript as model input during testing. The multiPA data was collected from real-world open-response scenarios. Detailed descriptions of the datasets are provided below.

\subsubsection{speechocean762}
The speechocean762 dataset contains 5,000 English utterances from 250 non-native speakers, each utterance labeled at the sentence, word, and phone level. We followed the training and testing split provided by the dataset and focused solely on the sentence and word-level labels. The sentence-level labels include the accuracy, fluency, prosody, and total scores, whereas word-level labels consist of accuracy, stress, and total scores. The utterances in the dataset are from 2 to 20 seconds long. 

\subsubsection{multiPA data}

The multiPA dataset comprises 50 audio clips, each lasting 10 to 20 seconds, obtained from about 20 anonymous dialog chatbot users. These users were native Mandarin speakers who used the dialog chatbot to practice their English. As a real use case for automatic pronunciation assessment, users can access the system with their own headsets at their own location. We recruited five annotators with high proficiency in English to annotate the audio clips at both sentence and word levels. The sentence labels include accuracy, fluency, and prosody scores, graded on a scale from 1 (very poor) to 5 (excellent). At the word level, the focus was on intelligibility; annotators used four levels to mark segments of speech: (1) cannot understand, (2) challenging to understand but recognizable, (3) somewhat inaccurate but quite understandable, and (4) good. The target scores used for analysis were the average scores from these five annotators. We have obtained IRB approval to collect data, and we have undergone an ethics assessment.

\subsection{Model details and evaluation}
For all experiments, we train the model with a batch size of 2 and an SGD optimizer with learning rate 5e-5 and momentum 7e-1. All models are trained using early stopping, with a patience of 2. 10\% of the training data is used as the validation set for model selection and early stopping detection. We use the Pearson correlation coefficient (PCC) as the main evaluation metric because it has often been used in previous studies and provides better interpretability when comparing the performance on in-domain and out-of-domain data. If models cannot generate assessment scores (e.g., the forced aligner fails to align the text to speech), the lowest scores in the speechocean762 training data are used as alternatives. Lastly, we repeat each experiment five times with different random seeds and report the mean and standard deviation of the results. 

\section{Results}

\subsection{Model performance}\label{subsec:sys_performance}

\begin{table*}
\caption{Comparison of model performance. GT transcript free indicates, under the original design, whether the model was evaluated without using ground-truth transcripts.}
\renewcommand{\arraystretch}{1.15}
\centering
\resizebox{0.9\textwidth}{!}{%
\begin{tabular}{|c|c|ccc|cccc|}
\hline
\multirow{2}{*}{} &
  \multirow{2}{*}{\begin{tabular}[c]{@{}c@{}}GT transcript \\ free\end{tabular}} &
  \multicolumn{3}{c|}{Word-level score (PCC)} &
  \multicolumn{4}{c|}{Sentence-level score (PCC)} \\ \cline{3-9} 
 &
   &
  \multicolumn{1}{c|}{Accuracy} &
  \multicolumn{1}{c|}{Stress} &
  Total &
  \multicolumn{1}{c|}{Accuracy} &
  \multicolumn{1}{c|}{Fluency} &
  \multicolumn{1}{c|}{Prosody} &
  Total \\ \hline \hline
\begin{tabular}[c]{@{}c@{}}GOPT\\ (medium.en)\end{tabular} &
  \ding{55} &
  \multicolumn{1}{c|}{0.273} &
  \multicolumn{1}{c|}{0.067} &
  0.265 &
  \multicolumn{1}{c|}{0.528} &
  \multicolumn{1}{c|}{0.527} &
  \multicolumn{1}{c|}{0.545} &
  0.528 \\ \hline 
  
  \hline
Lin \textit{et al.}~\cite{lin2023exploiting} &
  \ding{51} &
  \multicolumn{1}{c|}{-} &
  \multicolumn{1}{c|}{-} &
  - &
  \multicolumn{1}{c|}{\textbf{0.725}} &
  \multicolumn{1}{c|}{-} &
  \multicolumn{1}{c|}{-} &
  - \\ \hline
Liu \textit{et al.}~\cite{liu2023zero} &
  \ding{51} &
  \multicolumn{1}{c|}{-} &
  \multicolumn{1}{c|}{-} &
  - &
  \multicolumn{1}{c|}{-} &
  \multicolumn{1}{c|}{-} &
  \multicolumn{1}{c|}{-} &
  0.60 \\ \hline
Liu \textit{et al.}~\cite{liu2023asr} &
  \ding{51} &
  \multicolumn{1}{c|}{-} &
  \multicolumn{1}{c|}{-} &
  - &
  \multicolumn{1}{c|}{-} &
  \multicolumn{1}{c|}{\textbf{0.795}} &
  \multicolumn{1}{c|}{-} &
  - \\ \hline
vanilla SSL &
  \ding{51} &
  \multicolumn{1}{c|}{-} &
  \multicolumn{1}{c|}{-} &
  - &
  \multicolumn{1}{c|}{\begin{tabular}[c]{@{}c@{}}0.692\\ (std:0.006)\end{tabular}} &
  \multicolumn{1}{c|}{\begin{tabular}[c]{@{}c@{}}0.757\\ (std:0.010)\end{tabular}} &
  \multicolumn{1}{c|}{\begin{tabular}[c]{@{}c@{}}0.757\\ (std:0.009)\end{tabular}} &
  \begin{tabular}[c]{@{}c@{}}0.714\\ (std: 0.006)\end{tabular} \\ \hline
  \rowcolor[HTML]{EFEFEF}
\textbf{MultiPA} &
  \ding{51} &
  \multicolumn{1}{c|}{\begin{tabular}[c]{@{}c@{}}\textbf{0.427}\\ (std: 0.008)\end{tabular}} &
  \multicolumn{1}{c|}{\begin{tabular}[c]{@{}c@{}}\textbf{0.239}\\ (std:0.025)\end{tabular}} &
  \begin{tabular}[c]{@{}c@{}}\textbf{0.436}\\ (std:0.010)\end{tabular} &
  \multicolumn{1}{c|}{\begin{tabular}[c]{@{}c@{}}0.705\\ (std:0.009)\end{tabular}} &
  \multicolumn{1}{c|}{\begin{tabular}[c]{@{}c@{}}0.772\\ (std:0.010)\end{tabular}} &
  \multicolumn{1}{c|}{\begin{tabular}[c]{@{}c@{}}\textbf{0.764}\\ (std:0.016)\end{tabular}} &
  \begin{tabular}[c]{@{}c@{}}\textbf{0.730}\\ (std:0.006)\end{tabular} \\ \hline 
\end{tabular}%
}
\label{tab:system_performance}
\end{table*}

Table~\ref{tab:system_performance} presents a comparison of model performance without using the ground-truth transcript. The term \emph{GT transcript free} indicates whether the original design evaluated the model under the assumption that ground-truth transcripts were unavailable. First, we conducted a comparison between our model and the GOP-based model in an open-response scenario. We selected GOPT~\cite{gong2022transformer} as a representative because it is the basis for several studies~\cite{do2023hierarchical, chao20223m, do2023score}, and its code is open-source. To use GOPT in the open-response scenario, we used the transcript from Whisper medium.en to replace the ground-truth transcript, following the MultiPA setting. The experimental results demonstrate that MultiPA outperforms GOPT significantly. Despite both models being trained with ground-truth transcripts, MultiPA's structure is more robust for handling inaccurate transcripts and can be directly applied in open-response scenarios. Note that, when evaluating in the open response scenario, there is a potential mismatch between ground-truth word labels and the assessed scores because the assessment is based on ASR-recognized words. Therefore, we used Charsiu's alignment information to force-align ground-truth words to each recognized word, and employed the average score of aligned ground-truth words as the target score for the corresponding recognized word. In this way, although the ASR-recognized words might be incorrect, both the target and predicted scores refer to similar portions of the speech signal. 

We compared MultiPA with models that were evaluated without using the ground-truth transcripts. Results show that our model achieved comparable or higher performance for the single task while providing more comprehensive assessment scores for different aspects. We also include vanilla SSL as one of the baselines, which fine-tuned a pre-trained SSL model by average-pooling the SSL’s output embeddings and adding a dense output layer for sentence-level scores. While vanilla SSL has the ability to provide sentence-level assessments, it lacks the capability to provide a word-level assessment due to the absence of information on word boundaries. With the additional features introduced in MultiPA, MultiPA is able to provide word-level assessment and more accurate sentence-level assessment. 

\subsection{Ablation study of ASR models} 

\begin{figure}[htb]
\centering
\centerline{\includegraphics[scale=0.87]{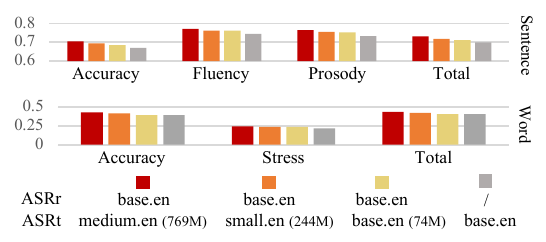}}
\caption{Ablation studies for using different ASR models.}
\label{fig:exp_asr}%
\end{figure}

We conducted ablation studies to show the performance impact of using different ASR models which differ mainly in model size. In Figure~\ref{fig:exp_asr}, we observed that, while ASRr is fixed as Whisper base.en, employing medium.en produces the highest scores, whereas utilizing base.en results in the lowest. This suggests that employing ASR models with greater diversity can enhance overall performance. In addition, the improved performance of models with ASRr compared to one without reflects the effectiveness of integrating ASRs and alignment features.
\subsection{Analysis of real-world open response scenario data} 
\begin{figure}[htb]
\centering
\centerline{\includegraphics[scale=0.95]{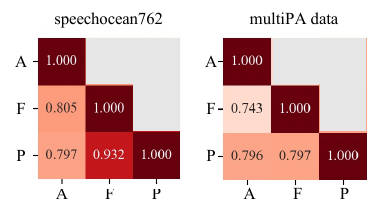}}
\caption{Correlation between different pronunciation tasks. \emph{A}, \emph{F}, \emph{P} refer to accuracy, fluency, and prosody, respectively.}
\label{fig:correlation}%
\end{figure}

We first calculated correlations between different pronunciation tasks (Figure~\ref{fig:correlation}) in both speechocean762 and multiPA data. We observed that, in the speechocean762 dataset, the correlation between fluency and prosody is higher compared to other task pairs. This strong correlation could be attributed to the similarity in the instructions for assessing fluency and prosody. For example, the criterion for the highest fluency score is \emph{“coherent speech, without noticeable pauses, repetition or stammering,”} and that for prosody is \emph{“correct intonation, stable speaking speed, and rhythm.”} Because the \emph{“stable speaking speed”} implies \emph{“without noticeable pauses, repetition or stammering”}, the resulting fluency and prosody scores could be very similar. To better differentiate between fluency and prosody, we revised the instruction when collecting multiPA data. For instance, we defined the highest fluency score as \emph{“fluent without noticeable pauses or stammering”}, following the setting of speechocean762, but modified the highest prosody score to \emph{“excellent prosody, expressive and well-modulated tone, enhancing communication with effective pitch and rhythm variations.”} As a result, in multiPA data, the correlation between prosody and fluency is lower than that in the speechocean762 dataset. 

We evaluated MultiPA on the real-world data (Table~\ref{tab:multiPA_data}). 
The results demonstrate MultiPA's ability to capture sentence-level accuracy and fluency proficiency, achieving correlations exceeding 0.6. However, we observed notably lower sentence-level prosody performance. This difference could be attributed to modifications made to the prosody assessment instructions during the collection of multiPA data, as previously discussed (Figure~\ref{fig:correlation}). We also examined the multitask learning approach by comparing the performance of the MultiPA structure with only sentence tasks (denoted as \emph{-sentence}) and with only word-level tasks (denoted as \emph{-word}). The experimental results show that multitask learning, incorporating both sentence-level and word-level assessment into the same model, can enhance performance compared to single-task learning. The word-level assessment is especially helpful for sentence-level accuracy because the word-level score mainly focuses on word accuracy. 

\begin{table}[htbp!]
\caption{Performance on multiPA data (PCC).}
\centering
\resizebox{0.94\linewidth}{!}{%
\begin{tabular}{|c|c|c|c|c|}
\hline
& Accuracy & Fluency & Prosody & Word  \\ \hline \hline
\rowcolor[HTML]{EFEFEF} 
MultiPA &
  \begin{tabular}[c]{@{}c@{}}\textbf{0.62}\\ (std:0.04)\end{tabular} &
  \begin{tabular}[c]{@{}c@{}}\textbf{0.65}\\ (std:0.05)\end{tabular} &
  \begin{tabular}[c]{@{}c@{}}\textbf{0.49}\\ (std:0.03)\end{tabular} &
  \begin{tabular}[c]{@{}c@{}}\textbf{0.39}\\ (std:0.03)\end{tabular} \\ 
  \hline 
\multicolumn{1}{|r|}{\textit{-sentence}} &
  \begin{tabular}[c]{@{}c@{}}0.57\\ (std:0.03)\end{tabular} &
  \begin{tabular}[c]{@{}c@{}}0.60\\ (std:0.04)\end{tabular} &
  \begin{tabular}[c]{@{}c@{}}0.48\\ (std:0.03)\end{tabular} &
  - \\ \hline
\multicolumn{1}{|r|}{\textit{-word}} & -        & -       & -       & \begin{tabular}[c]{@{}c@{}}0.37\\ (0.02)\end{tabular}                                            \\ \hline
\end{tabular}%
}
\label{tab:multiPA_data}
\end{table}

Our analysis reveals limitations in the current model's ability to evaluate word-level scores, potentially due to the highly imbalanced word-level labels in the speechocean762 dataset~\cite{do2023score}. To address this, we proposed a strategy that can leverage our current model effectively in real-world scenarios. We re-framed the word-level assessment from measuring the “accuracy score of a word's pronunciation” to a binary mispronunciation detection~\cite{ryu2023joint}: “whether the pronunciation of the word needs improvement.” We defined “a word that needs improvement” as a word that does not receive the highest score from all annotators. Because L2 learners might be overwhelmed by too many suggestions, we care about \emph{precision} (i.e., whether the words suggested by the model actually need improvement) rather than recall (whether the model captured all the words that need improvement.) By setting the threshold less than \emph{mean-std}, the model can achieve over 0.9 precision in word suggestions. This means that, if the model suggests a word with a score significantly lower than the scores from all data, then 
the suggested word highly likely needs improvement. Specifically, the precision is 0.926 (213/230 words) when setting the threshold as \emph{mean-std}, and the baseline precision of suggesting all words is 0.759 (1092/1439 words). The threshold can be customized based on the pronunciation levels of all users or the sensitivity level at which the user desires feedback from the model.


\section{Conclusion}
\label{sec:conclusion}
We introduce MultiPA, a multi-task speech pronunciation assessment model that provides comprehensive assessment at both the sentence and word levels. Our study includes the collection of pilot data from real-world use cases, model evaluation on out-of-domain data, and analysis of correlations between different pronunciation tasks. The pilot data collected in this study will be made available for other researchers to evaluate their pronunciation assessment models. One limitation of MultiPA is that the assessed word-level scores in the open response scenario correspond to ASR-recognized words, but these words might be incorrect and vary between different runs. With the time alignment information of these words, MultiPA can still indicate specific parts of the speech signal that need improvement; however, how such information can effectively help learners to improve their pronunciation requires further investigation. In future, we plan to explore data augmentation or self-supervised methods and collect feedback from the model users to improve the model’s performance further.

\section{Acknowledgements}
The authors would like to thank Label Studio\footnote{\url{https://labelstud.io/}} for supporting researchers and making data collection more accessible. This work was partially funded by the NIH National Institute on Aging under GRANT 5 R01AG081928-02.


\bibliographystyle{IEEEtran}
\bibliography{mybib}

\end{document}